\def\year{2020}\relax
\documentclass[letterpaper]{article} % DO NOT CHANGE THIS
\usepackage{aaai20}  % DO NOT CHANGE THIS
\usepackage{times}  % DO NOT CHANGE THIS
\usepackage{helvet} % DO NOT CHANGE THIS
\usepackage{courier}  % DO NOT CHANGE THIS
\usepackage[hyphens]{url}  % DO NOT CHANGE THIS
\usepackage{graphicx} % DO NOT CHANGE THIS
\usepackage{graphicx}  % DO NOT CHANGE THIS
\usepackage{amssymb}
\usepackage{amsmath}
\usepackage{mathtools}
\usepackage[dvipsnames]{xcolor}

\title{Using Meta-Knowledge Mined from Identifiers to \linebreak Improve Intent Recognition in Neuro-Symbolic Algorithms}
\author{\large \textbf{Claudio Pinhanez, Paulo Cavalin, Victor Ribeiro, Heloisa Candello, Julio Nogima, Ana Appel,} \\ \large \textbf{Mauro Pichiliani, Maira Gatti de Bayser, Melina Guerra, Henrique Ferreira, Gabriel Malfatti} \\ \\ \Large IBM Research - Brazil}

\usepackage[switch]{lineno}  %

\begin{document}
%\linenumbers  %
\maketitle

\begin{abstract}
In this paper we explore the use of meta-knowledge embedded in intent identifiers to improve intent recognition in conversational systems. As evidenced by the analysis of thousands of real-world chatbots and in interviews with professional chatbot curators, developers and domain experts tend to organize the set of chatbot intents by identifying them using proto-taxonomies, i.e., meta-knowledge connecting high-level, symbolic concepts shared across different intents. By using neuro-symbolic algorithms able to incorporate such proto-taxonomies to expand intent representation, we show that such mined meta-knowledge can improve accuracy in intent recognition. In a dataset with intents and example utterances from hundreds of professional chatbots, we saw improvements of more than 10\% in the equal error rate (EER) in almost a third of the chatbots when we apply those algorithms in comparison to a baseline of the same algorithms without the meta-knowledge. The meta-knowledge proved to be even more relevant in detecting out-of-scope utterances, decreasing the false acceptance rate (FAR) in more than 20\% in about half of the chatbots. The experiments demonstrate that such symbolic meta-knowledge structures can be effectively mined and used by neuro-symbolic algorithms, apparently by incorporating into the learning process higher-level structures of the problem being solved. Based on these results, we also discuss how the use of mined meta-knowledge can be an answer for the challenge of knowledge acquisition in neuro-symbolic algorithms.

%In this paper we explore the use of meta-knowledge embedded in intent identifiers to improve intent recognition. As we show in the analysis of thousands of real-world chatbots, developers tend to organize the set of intents by making use of proto-taxonomies, i.e. meta-knowledge including high-level concepts shared across different intents. By means of approaches that are able to use the proto-taxonomies to expand intent representation, we show that such mined meta-knowledge can be useful to improve intent recognition, specially in detecting out-of-scope examples. In about 50\% of the chatbots there is at least 20\% decrease in false acceptance rates. In summary, this work demonstrates that chatbot developers tend to use high-level knowledge, and such knowledge can be effectively used in a neuro-symbolic setting to learn the high-level language of the developers. 

\end{abstract}

\section{Introduction}

After almost a decade of notable advances in AI using data-driven machine learning approaches, there is a growing sense in the field that symbolic knowledge needs to be included in AI systems to get to the next level of machine intelligence. This thought is materialized in the so called \textit{Neuro-Symbolic} approaches which have already produced some intriguing results~\cite{parisotto2017neurosymbolic,besold2017neuralsymbolic,Tenenbaum2011,bengio2017consciousness,Mao2019NeuroSymbolic,hudson2019learning,raedtetal2019}.

However, even if successful, such approaches will require symbolic data or knowledge to be captured and represented for the machine. Eliciting knowledge directly from human beings has been proved to be a difficult task, both in the cases of  specialized, professional knowledge~\cite{boose1989survey} and common-sense~\cite{davis2014representations,singh2002open}. Similarly, mining reliably symbolic knowledge from text sources still remains a difficult task, in spite of many advancements in mining of knowledge graphs
~\cite{fossati2015,ji2020survey,asim2018survey}. As much as collecting large amounts of reliable data has become a bottle-neck for the use of data-driven machine learning, getting knowledge from textual data or human beings into reasonably complete and correct symbolic representations is likely to be a major issue for neuro-symbolic methods.

This paper explores an alternative path which explores meta-knowledge developers of AI systems sometimes embed in their source code. In particular, we examine the case of professional conversational systems and the symbolic knowledge their developers often embed in identifiers. Analyzing a dataset comprising thousands of different conversational systems developed in a very popular platform, we observe a very common pattern of using a symbolic structure, called here \textit{proto-taxonomy}, to name the intents to be recognized by the system. This practice was also verified qualitatively in workshops we conducted with developers.

In spite many consistency and incompleteness issues with those proto-taxonomies, we show that they can be employed to improve the accuracy of recognition, using adaptations of recent neuro-symbolic methods. 
%We also demonstrate that the embedded meta-knowledge can be used for zero-shot learning, in particular for the task of mining new user intents from call-logs of conversations. In the latter case, the mined clusters of intents not only conform to the overall structure of the proto-taxonomies, but also are tentatively named by the system following the developers' patterns. In other words, the proto-taxonomies can be used to make the machine "think" and "act" following the mental models of the developers of the systems. We argue in the paper that, by doing so, the proposed methods also add explainability to the AI system.
This seems to signal towards neuro-symbolic techniques designed to handle imperfect knowledge representations. We see as a needed compromise to bring back symbolic reasoning to AI in a sustainable form, avoiding the old pitfalls of \textit{knowledge engineering}~\cite{hayes1984industrialization,studer1998knowledge,studer1999knowledge,chang2001handbook}.

This paper starts by looking into the recent advances in neuro-symbolic systems, reviewing the difficulties in knowledge mining, and exploring previous use of informal knowledge. We then present the evidence found that developers of conversational systems embed meta-knowledge within the source code of their systems. We follow by describing algorithms integrating such meta-knowledge into intent recognition algorithms and by evaluating them first with two typical intent recognition datasets, and then with hundreds of workspaces created in a professional tool called here \textit{ChatWorks}. The results show most of those workspaces can benefit from the techniques described in this paper.

\section{Related Work}
Neuro-symbolic approaches combine statistical methods with logic symbolism:  ``neural-symbolic systems aim to transfer principles and mechanisms between (often nonclassical) logic-based computation and neural computation''~\cite{besold2017neuralsymbolic}. Such kind of systems are viewed as a way to embed high-level knowledge and even some form of ``consciousness'' into machine learning systems, making the language to develop them closer to ``what passes in a man's own mind''
~\cite{bengio2017consciousness}, which would likely make those systems more explainable than current deep learning algorithms.

%\subsection{Neuro-Symbolic Methods and Techniques}
Although neuro-symbolic systems are not new, we observe increasing interest on this approach in recent years, resulting in a myriad of novel techniques applied to different problems, contexts, and scenarios~\cite{parisotto2017neurosymbolic,manhaeve2018,garcez2019neuralsymbolic,hudsonmanning2019,raedtetal2019}. For instance, in \cite{Mao2019NeuroSymbolic}, an approach for image understanding is suggested which takes the object-based scene representations and translates sentences into executable, symbolic programs. In~\cite{oltramari2020neurosymbolic}, embeddings computed from knowledge graphs are used as attention layers for tasks such as autonomous driving (AV) and question-answering. And in \cite{kartsaklis2018mapping}, embeddings from a knowledge graph are mapped to sentence embeddings for tasks such as {\it the inverse dictionary problem}.

%\subsection{Knowledge Acquisition and Mining from Experts}
One important requirement for many neuro-symbolic systems is to represent knowledge in a structured format such as knowledge graphs, ontologies, or taxonomies \cite{ji2020survey}. In some cases, such as the scene ontology for AV in \cite{oltramari2020neurosymbolic}, a lot of effort was needed to be put on manual annotation. Nevertheless, as presented in \cite{fossati2015}, an unsupervised approach can sometimes be used to mine the meta-knowledge introduced by the experts, such as the categories in Wikipedia pages.

The high-level representation in intent identifiers can be viewed as similar to comments included in programming source codes. Code commentaries are one of the means employed by developers to help them organize their thought process while producing code.
%This practice results in code documentation that is valuable not only to those who produce them but also to anyone trying to understand source code organization and logic.
Research on code commentaries has shown that they can be useful for automatic generation of code, consistency check, classification and quality evaluation \cite{yang2019}. Similar behavior of users for organizing content can also be observed in e-mails \cite{Whittaker2011}, computer files \cite{barreau1995,jones2005,civan2008}, and \textit{Jupyter} notebooks~\cite{rule2018ten}.

Considering our  context of intent recognition, intent identifiers might contain a high-level representation of the main content of the intent. As shown in~\cite{Chen2016}, intent identifiers can be formatted as natural language sentences to learn a model which maps training examples into those sentences, so that the meta-knowledge can be used in \textit{zero-shot learning}
~\cite{wang2019survey}. Unfortunately, the dataset explored in this work is very limited. Recent work has also demonstrated that intent recognition can be improved with enhanced class representations such as \textit{word-graphs}~\cite{cavalin2020improving} by mining symbolic knowledge from the example utterances.
%\subsection{Embedded Meta-Knowledge from Comments}

This work aims to fill in some of those gaps by providing a better understanding of the usefulness of the meta-knowledge embedded in intent identifiers by exploring a large set of intent recognition datasets; and by going deeper into the symbolic representations of the identifiers viewing them as quasi taxonomies.

\section{Embedded Meta-Knowledge in Intents of Conversational Systems}

Most real-world, deployed conversational systems in use today have been built based on the rule-based \textit{intent-action} paradigm, using platforms such as \textit{Luis.ai}, \textit{Watson Assistant}, or \textit{Alexa Skills}. Each \textit{intent} corresponds to a desired information or answer from the user and is defined by a set of exemplar utterances by the chatbot developers. During runtime, each utterance from the user is recognized as one of the defined intents or as \textit{out-of-scope} (OOS), and the associated action is generated, often a pre-written sentence created by developers or subject-matter experts (SMEs).

\begin{figure}[t!]
  %\begin{minipage}{\marginparwidth}
    \centering
    \includegraphics[trim=0cm 0.5cm 0cm 0cm,width=4cm]{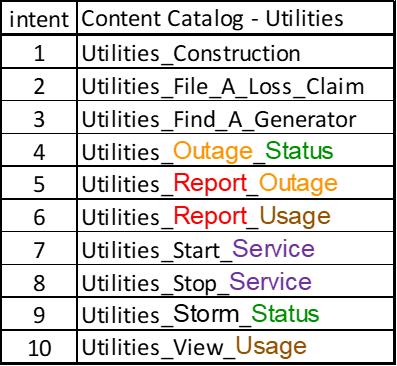}
    \caption{Pre-defined intents for utilities-related chatbots of the \textit{Watson Assistant}     platform.}
    \label{fig:intents_utilities.png}
  %\end{minipage}
\end{figure}

\begin{figure}[t!]
  %\begin{minipage}{\marginparwidth}
    \centering
    \includegraphics[trim=0cm 0.5cm 0cm 0cm,width=6cm]{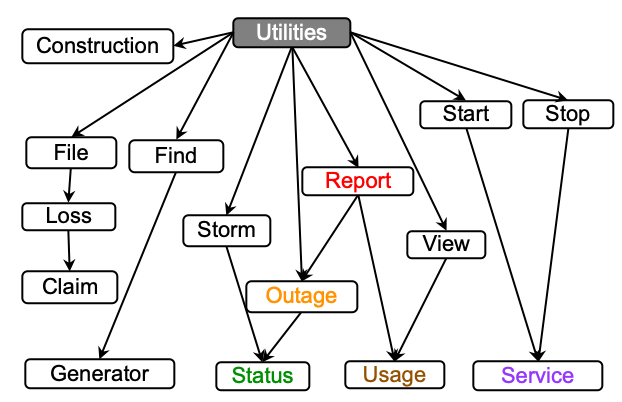}
    \caption{The intent proto-taxonomy associated to the utilities-related intents of fig.~\ref{fig:intents_utilities.png}.}
    \label{fig:intents_utilities_graph}
  %\end{minipage}
\end{figure}

Many of those platforms also come with a pre-defined, domain-specific list of intents which can be added to any chatbot to speed-up development. For example, fig.~\ref{fig:intents_utilities.png} shows the list of pre-defined intents from \textit{Watson Assistant} for utilities-related conversational systems. Notice that the names of intents aim to describe the meaning of each pre-defined intent by representing it through a sequence of keywords separated by underscore characters. Some of those keywords appear many times (marked in colors), with a structure which has semblance to a \textit{taxonomy}.

This pattern of naming intents following a categorical path can also be found in the pre-defined intents of other platforms and, as we show in this paper, in the names of the intents defined by developers themselves. The goal seems to provide the intent classes with a summarized description of each intent in a way that the similarity of different intents is highlighted. Such patterns are also common in the way people organize files and e-mails in computers \cite{civan2008,Whittaker2011} and how software developers name functions \cite{yang2019}.

%\subsection{Intent Proto-Taxonomies}

At the same time, by regarding the keywords in the name intents as basic concepts and the underscore characters as a connection between them, we can structure the list of intent identifiers as a sort of very basic knowledge graph~\cite{ehrlinger2016towards}, here referred as \textit{intent proto-taxonomies}. Figure~\ref{fig:intents_utilities_graph} depicts the intent proto-taxonomy associated to the list of the intents in fig.~\ref{fig:intents_utilities.png}.

% {\bf COMMENT Paulo: Talvez removeria o parágrafo abaixo. Ou manteria a primeira frase e condensaria com o proximo}

% A basic inspection of the example of intent proto-taxonomy shown in fig.~\ref{fig:intents_utilities_graph} reveals that, as a knowledge graph, it has several shortcomings. For instance, we would expect, by analogy, that the intent 1, 4, and 8, would have an action verb as its second keyword like all the others, possibly the keyword ``Report''. Similarly, the concept of ``Service'' most likely can be checked for its status, perhaps adding a couple of links to enable an intent named ``Report\_Service\_Status''. 

% What we see is that, even in this case of a professional list of intents provided by a highly developed tool, the meta-knowledge mined from the list of the intent names is going to have inconsistencies and be incomplete. However, it has two great qualities: (1) it is embedded in the conversational system by its developers, so there is no need of knowledge acquisition from experts; (2) it is easily mined.

A basic inspection of the intent proto-taxonomy shown in fig.~\ref{fig:intents_utilities_graph} reveals that, as a knowledge graph, it has several shortcomings, such as the lack of action verbs in some intents. In addition, even in this case of a professional list of intents provided by a highly developed tool, the meta-knowledge mined from the list of the intent identifiers has inconsistencies and seems to be incomplete. However, it has two great qualities: (1) it is embedded in the conversational system, so there is no need of knowledge acquisition from experts; (2) it is easily mined.

The key question is whether, given its limitations as discussed above, the embedded knowledge is good enough to be used by neuro-symbolic algorithms. We will show later that, indeed, this meta-knowledge can enhance machine learning algorithms. But first let us examine the evidences we found that the practice of naming intents in conversational systems using an intent proto-taxonomy is a fairly common practice, and thus able to provide structured domain meta-knowledge almost ``for free'' for a large number of professional systems in use today.

\subsection{How Developers Use Intent Proto-Taxonomies}

We conducted a 4-day design workshop with four expert developers to understand the challenges SMEs and developers of conversational systems have and what could facilitate their work \cite{chi2021paper}. Those SMEs have developed chatbots for the auto industry, banking, and telco using the ChatWorks platform (anonymized for review).

% We conducted a 4-day design workshop to understand the challenges SMEs and developers of conversational systems have and what could facilitate their work. Four expert developers were invited to participate, joined by 4 of the authors of this paper. The SMEs have developed chatbots for the auto industry, banking, and telco, always using the ChatWorks platform. The workshop and its results are described in detail in~\cite{chi2021paper}.

% \begin{figure}[t!]
%   %\begin{minipage}{\marginparwidth}
%     \centering
%     \includegraphics[width=9cm]{figures/schema_taxonomy_cars.png}
%     \caption{Intent taxonomy of a car chatbot as structured by the expert curators in a kind of \textit{mindmap} diagram.}
%     \label{fig:taxonomy_drawing.png}
%   %\end{minipage}
% \end{figure}

The structuring of the intent identifiers using proto-taxonomies was discussed and explored with them. The SMEs reported a very pro-active practice of naming intents following a formally defined structure, typed in a sort of taxonomy, and shared among their peers and domain experts in the clients, who were also responsible for maintaining those systems. Some of them brought \textit{mindmaps} to explain the concept relations in the \textit{workspace} (the set of all intents) %~\ref{fig:taxonomy_drawing.png} 
and showed how they make available those concepts to their team in the system or using spreadsheets. They told us that often underscore characters are used to separate concepts and that the order of the concepts usually represents how the workspace is organized.

The \textit{taxonomy}, as they often refer to it, is also a kind of self-indexing information for future use, a name that, by representing the semantics of an intent, can be used to simplify their work and collaboration. This study provided evidence that the use of structured meta-knowledge in the intent identifiers was an intentional and well-established practice among some developers. The remaining question was how widespread this practice was. 

\subsection{The Use of Intent Proto-Taxonomies in ChatWorks}

The ChatWorks platform has an opt-in feature in which developers of chatbots can share their code and content (called \textit{workspaces}) with the company which owns the platform for research and development purposes. We were given access to about 18K workspaces active in six months between 2019 and 2020, all of them in English language.
%In this study, the 26K were filtered to remove duplicates and workspaces with less than 8~intents. The resulting dataset is composed of 6,735 workspaces (3,840 for English and 2,895 for Portuguese). About 81\% of English and Portuguese workspaces had from 8 to 100 intents, and the biggest workspaces had 1,974 intents for English and 1,842 for Portuguese. 
Those workspaces were filtered to remove duplicates and workspaces with less than 8~intents. The resulting dataset is composed of 3,840 workspaces. About 81\% of them had from 8 to 100 intents and the largest had 1,974 intents. 

%Using two different methods, namely regular expressions and perplexity, we processed all workspaces to determine the most efficient separator for the keywords, if it existed. 

We used two criteria for taxonomy identification and size in each workspace: (1) the intent identifier must have a concept structure (words) separated by a symbol (separator); and (2) the concepts in the same position should be able to group themselves at least in two different classes. Given a workspace with a set of intent identifiers, we first ranked the best separator (period, underline, camelcase, or dash) to split the name into concepts.
%The ranking method was based first on an exhaustive search of a commonly used separator (period, underline, camelcase, and dash). 
%We found widespread use of such separators and the most used separator for both languages was the underscore (75.9\% for English and 59.3\% for Portuguese), followed by camelcase (17.0\% and 26.7\%), dash (4.9\% and 9.0\%), and period (1.2\% and 4.4\%).

To this end, we calculated the \textit{perplexity}~\cite{manning1999foundations} of a bag of concepts using each separator and selected the one with minimum perplexity.
%Perplexity is a measure of uncertainty for a certain sequence of words (or concepts) appearing in a language model 
%For that, we built language models and computed the average perplexity using a standard leave-one-intent-out evaluation scheme. %, where language models are built with all intents but one, which is used to compute the metric.
% The average perplexity for all intents forms our final metric.
%The separator, which minimized perplexity, was chosen as the separator for those workspaces. 
Next, each intent identifier was split using the selected separator and the resulting list of intent identifiers was compared to each other by the concepts at the same level. In a level, if the concepts were either all equal or all different, then that level was not evaluated. When the grouping of concepts was possible,
%(i.e., at least two different concepts were repeating at the same level)
the intents with those concepts were selected as \textit{intents with taxonomy}. The \textit{taxonomy rate} was calculated by the ratio between the number of intents with taxonomy and the number of intents created by the user (excluding all the pre-defined domain-specific intents provided by ChatWorks).

%We also calculated the \textit{average depth} of the taxonomy as the average of the number of concepts of each intent with taxonomy. Also, the number of \textit{unique concepts} were calculated as an indicator of amount of concepts present in that workspace.

%Using those metrics, 76\% of the English workspaces, respectively 80\% for Portuguese, had a taxonomy rate above 10\%. Almost 52\% of all workspaces, in English, respectively 38\% in Portuguese, had a taxonomy rate above 50\%. Moreover, 16\% and 20\% of the workspaces, respectively had a very high taxonomy rate from 90\% to 100\%. 

\begin{figure}[t!]
    \centering
    \includegraphics[trim=1cm 1cm 1cm 1cm,width=\columnwidth]{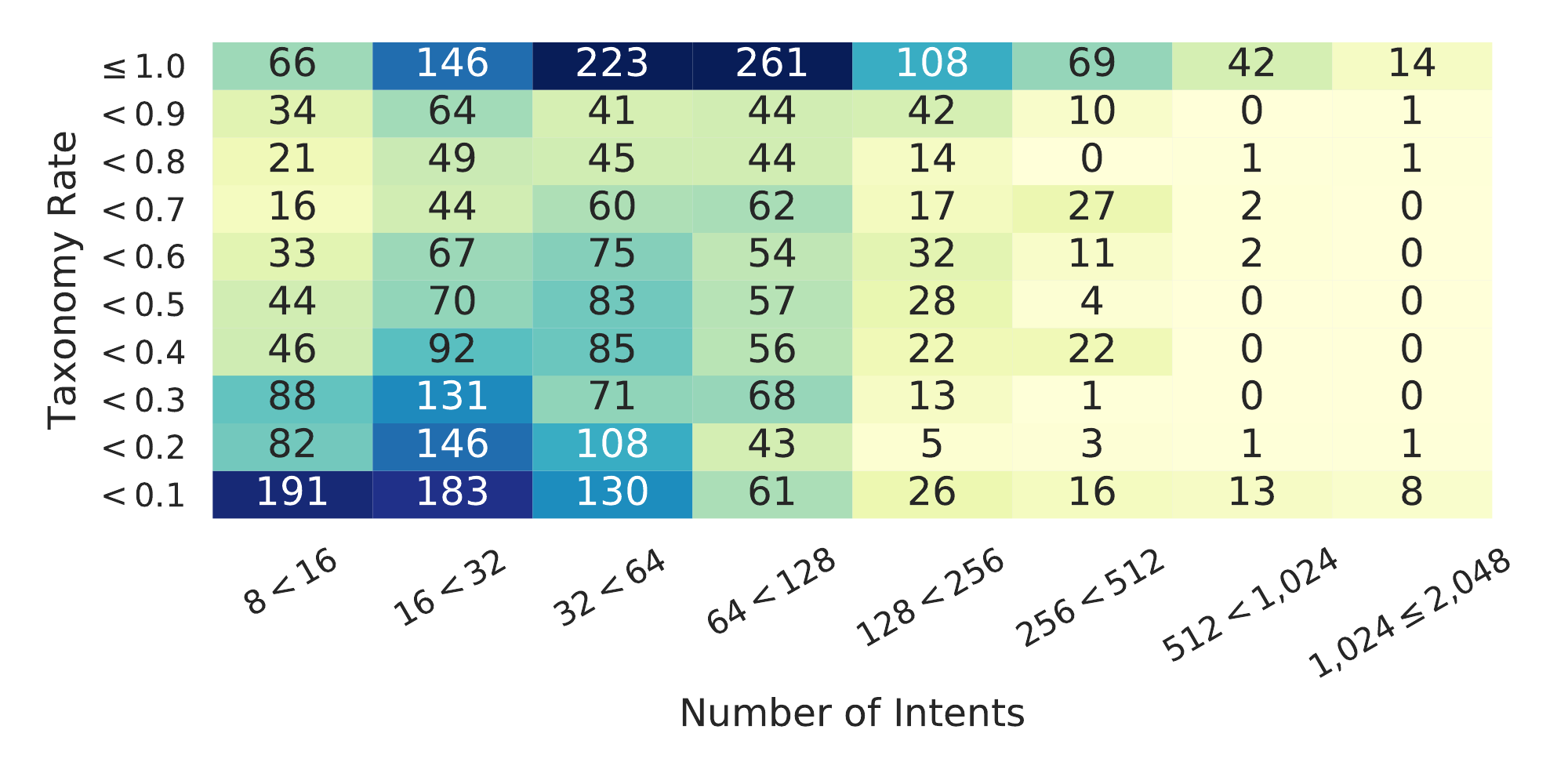}

    \caption{Distribution of the number of workspaces according to the taxonomy rate of the 3,840 English % (top) and 2,895 Portuguese (bottom) 
    workspaces. %Notice that the x axix is in logarithmic scale.
    }
    \label{fig:workspaces_dist}
\end{figure}

Using those metrics, 76\% of all 3,840 workspaces had a taxonomy rate above 10\%, almost 52\% had a taxonomy rate above 50\%, and 16\% had a very high taxonomy rate from 90\% to 100\%. 
%approximately uniformly among the other 10\% taxonomy ranges. 
Figure~\ref{fig:workspaces_dist} shows how the 3,840 workspaces are distributed considering both the total number of intents ($x$ axis) and the taxonomy rate ($y$ axis). Notice that the distribution follows a sort of ``step'' function where, as the threshold of 64 intents in the workspace is crossed, the majority of the workspaces had more than 50\% of taxonomy rate. It seems that, as the complexity of the workspace increases with the number of intents, more often developers and SMEs resort to structure the intents as a proto-taxonomy.

Notice that the same inconsistencies which were seen in the pre-defined intents from Watson Assistant of fig.~\ref{fig:intents_utilities.png} seem to be also present in the developers' workspaces. Nevertheless, the results of this analysis seem to overwhelming confirm that using intent proto-taxonomies is a fairly common practice in ChatWorks, reaching at least around 80\% of all workspaces and even more common in the workspaces with a high number of intents. 
%The results of this analysis seem to overwhelming confirm that using intent proto-taxonomies is a fairly common practice in ChatWorks, reaching at least 80\% of all workspaces, and even more common in the workspaces with a high number of intents. 

\section{Using Mined Meta-Knowledge to Improve Intent Recognition}
This section presents a formal description of the methodology employed in this work to take advantage of the proto-taxonomies in a neuro-symbolic approach.
% First we will describe how the set of classes has been expanded to make use of word graph information, inspired by a method proposed by \citet{kartsaklis2018mapping}, and then we describe how the method is used to detect OOS samples.
  
\subsection{Embedding the Set of Classes}

An \emph{intent classification} method is a function $D$ which maps a set of sentences (potentially infinite) $S=\{s_{1},s_{2},...\}$ into a finite set of classes $\Omega=\{\omega_{1},\omega_{2},...,\omega_{c}\}$:
\begin{equation}
    D:S\rightarrow\Omega \hspace{5mm} D(s)=\omega_{i} %\hspace{5mm}  s \in S, i=1, 2, ..., c
\end{equation}
 
To enable a numeric, easier handling of the input text, an embedding $\xi:S\rightarrow\mathbb{R}^{n}$ is often used, mapping the space of sentences $S$ into a vector space  $\mathbb{R}^{n}$, and defining a classification function $E:\mathbb{R}^{n}\rightarrow\Omega$ such as $D(s)=E(\xi(s))$. 
%In typical intent classifiers, $E$ is usually a \emph{softmax} function.
In typical intent classifiers, $E$ is usually composed of a function $C$ which computes the probability of $s$ being in a given class, followed by the \emph{arg max} function. In many intent classifiers, $C$ is the \emph{softmax} function.
%\begin{equation}
%S \overset{\xi}{\rightarrow} \mathbb{R}^{n} %\overset{E}{\rightarrow} \Omega 
%\end{equation}
\begin{equation}
S \overset{\xi}{\rightarrow} \mathbb{R}^{n} \overset{C}{\rightarrow} \mathbb{R}^{c} \overset{arg max}{\rightarrow} \Omega 
\end{equation}
 
This paper explores how to use embeddings in the other side of the classification functions, that is, by embedding the set $\Omega$ of classes into another vector space $\mathbb{R}^{m}$. The idea is to use class embedding functions which somehow capture the intent proto-taxonomies,
%from the training sets,
as we will show later. Formally, we use a \emph{class embedding} function $\psi:\Omega\rightarrow \mathbb{R}^{m}$, its inverse $\psi^{-1}$, and a function $M:\mathbb{R}^{n}\rightarrow\mathbb{R}^{m}$ to map the two vector spaces so $ D(s)=\psi^{-1}(M(\xi(s)))$.

\begin{equation}\label{eq:main} S \overset{\xi}{\rightarrow} \mathbb{R}^{n} \overset{M}{\rightarrow} \mathbb{R}^{m} \overset{\psi^{-1}}{\rightarrow} \Omega 
\end{equation}

In our work we use typical sentence embedding methods to implement $\xi$. To approximately construct the function $M$ we employ a basic \emph{Mean Square Error} (MSE) method using the training set composed of sentence examples for each class $\omega_{i} \in \Omega$.
%$Z=\{z_{1}, z_{2}, ... , z_{p}\}, z_{j}\in S$ which are assigned to a correspondent class by the function $l:Z\rightarrow\Omega$. 
%As we will see next, the training set can be used to construct the embedding function for the set of classes $\psi$ and an approximation for its inverse $\psi^{-1}$. In addition, typical pre-trained general-purpose sentence embeddings can also be used for that.
 
\subsection{Adapting \emph{Kartsaklis} %et al.'s 
Method (LSTM)}

Our algorithms are inspired by a text classification method proposed for the inverse dictionary problem, where text definitions of terms are mapped to the term they define, proposed in \cite{kartsaklis2018mapping}. 
The embedding of the class set into the continuous vector space (equivalent to the $\psi$ function in equation~\ref{eq:main}) is done by expanding the knowledge graph of the dictionary words with  nodes corresponding to words related to those terms and performing random walks on the graph to compute graph embeddings related to each dictionary node, using the \emph{DeepWalk} algorithm~\cite{perozzi2014deepwalk}. DeepWalk is a two-way function mapping nodes into vectors and back.

A \emph{Long Short-term Memory} (LSTM) neural network, composed of two layers and an attention mechanism, is used in~\cite{kartsaklis2018mapping} for mapping the input texts to the output vector space. 
To map the two continuous vector spaces representing the definition texts and the dictionary terms, a MSE function, learned from the training dataset, is used. 
%This approach achieves SOTA results on the reverse dictionary task and also in other tasks such as document classification and text-to-entity mapping.

%In this work, the approach from \cite{kartsaklis2018mapping} is employed for mapping the classes into a vector space, although we do not use a knowledge graph as described later. 
For this work, the knowledge graph is replaced by a \emph{proto-taxonomy graph} $G$ which associates each class to a node and connects to each of them nodes that correspond to meta-knowledge concepts related to each class. To better capture the sequential aspect of the proto-taxonomies, we also connect each class node to bigrams of concepts, i.e., the concatenation of two subsequent concepts. We represent this by the function $\zeta$, such as $\zeta (\Omega)=G$, which is also invertible. Substituting this in equation~\ref{eq:main},
%The classification model maintains the attention mechanisms and the two layer LSTMs. 
\begin{equation}\label{eq:ltsmoriginal}
S \overset{LSTM}{\rightarrow} \mathbb{R}^{n} \overset{MSE}{\rightarrow} \mathbb{R}^{m} \overset{DeepWalk^{-1}}{\rightarrow} G \overset{\zeta^{-1}}{\rightarrow} \Omega 
\end{equation}

In practice, we compute the mapping from the class embedding space into the class set, called here $InvG:\mathbb{R}^{m} \rightarrow \Omega$, simply by computing the distance $d$ between a point in $\mathbb{R}^{m}$ and the inverted projection of each class from $\Omega$ and then considering the closest class. That is, for each $w_{i} \in \Omega$, we consider the associated node in $G$ and compute the mapping in $\mathbb{R}^{m}$ of that node, as shown here:
\begin{equation} \small
InvG (x) = \underset{w_{i}}{\arg\min} \hspace{1mm} d(x,DeepWalk(G(w_i))
\end{equation}

By substituting this function into equation~\ref{eq:ltsmoriginal}, we obtain the algorithm we call here \emph{LSTM+T}:
\begin{equation}\label{eq:ltsmplust}
S \overset{LSTM}{\rightarrow} \mathbb{R}^{n} \overset{MSE}{\rightarrow} \mathbb{R}^{m} \overset{InvG}{\rightarrow} \Omega 
\end{equation}

For comparison, the traditional corresponding classification method is tested, where the graph embedding and associated functions are replaced by discrete \emph{softmax} outputs. We call this simply  \emph{LSTM}:
\begin{equation}
S \overset{LSTM}{\rightarrow} \mathbb{R}^{n} \overset{softmax}{\rightarrow} \mathbb{R}^{c} \overset{arg max}{\rightarrow} \Omega 
\end{equation}

\subsection{Replacing the LSTM with USE}

% The natural language processing community has been recently focusing attention on the novel \emph{transformer} models~\cite{vaswani2017attention}.
% This is due to the great performance improvement in several complex tasks, such as machine translation, question answering, and text classification.
% Moreover, such a performance is achieved without the use of convolutions or recurrence in neural networks.
% By using only the attention mechanism, models are built with lower computational costs, enabling the rapid development of larger and stronger general-purpose language models, which have been achieving SOTA performance in many different tasks, such semantic sentence similarity, text classification, just to name a few.

Recently, several general-purpose language models that can be used for computing sentence embeddings have been proposed, and the \emph{Universal Sentence Encoder} (USE) is one of them \cite{cer-etal-2018-universal}. Such an approach consists of a \emph{Transformer} neural network \cite{vaswani2017attention}, trained on varied sources of data, such as Wikipedia,  web news, web question-answer  pages  and  discussion forums. USE has achieved state-of-the-art results in various tasks, so we decided to try in our experiments as an alternative to the LSTM for the embedding of input sentences.
%That approach has been designed not only to serve as a baseline model to take advantage of transfer learning when little data is available, but also as a means to encode textual information, i.e., sentences, into real-valued $N$-dimensional embedding vector.

In this work we employed the multilingual USE version~3\footnote{https://tfhub.dev/google/universal-sentence-encoder-multilingual/3}.
By replacing LSTM with USE in eq.~\ref{eq:ltsmplust} we obtain algorithm \emph{USE+T}:
\begin{equation}
S \overset{USE}{\rightarrow} \mathbb{R}^{n} \overset{MSE}{\rightarrow} \mathbb{R}^{m} \overset{InvG}{\rightarrow} \Omega 
%\overset{DeepWalk^{-1}}{\rightarrow} G %\overset{\zeta^{-1}}{\rightarrow} \Omega 
%\hspace{0.5cm} s \in S, w_{i} \in \Omega 
\end{equation}

Like in the previous case, we also use the USE algorithm with traditional discrete softmax outputs for comparison, called here \emph{USE}:
\begin{equation}
S \overset{USE}{\rightarrow} \mathbb{R}^{n} \overset{softmax}{\rightarrow} \mathbb{R}^{c} \overset{arg max}{\rightarrow} \Omega 
\end{equation}

\vspace{0.1mm}

\subsection{Replacing DeepWalk with USE and CDSSM}
To explore variants of algorithms for embedding the classes and also approaches which do not need to be trained from scratch and allow on-the-fly handling of meta-knowledge, we tried replacing DeepWalk with two different methods.

The first one consists of applying USE sentence embeddings also for class embeddings, such as in eq.~\ref{eq:deepwalkreplacement}. To simplify notation, \emph{EMB} represents either LSTM or USE embeddings for the input text.
\begin{equation}\label{eq:deepwalkreplacement}
S \overset{EMB}{\rightarrow} \mathbb{R}^{n} \overset{MSE}{\rightarrow} \mathbb{R}^{m} \overset{USE^{-1}}{\rightarrow} G \overset{\zeta^{-1}}{\rightarrow} \Omega 
\end{equation}

This approach is similar to the way DeepWalk works but instead of training the graph embeddings from scratch, the class embeddings are represented by the mean sentence embedding computed from different random walks starting in the class node. We name these methods \emph{LSTM+S} and \emph{USE+S}, for EMB set with LSTM and USE, respectively.

%\subsection{Replacing the DeepWalk with CDSSM}
Additionally, we also evaluate the replacement of DeepWalk by the \emph{Convolutional Deep Structured Semantic Model} (CDSSM) proposed in \cite{Chen2016}, yielding the following algorithm where \emph{EMB} can be either LSTM or USE embeddings.
\begin{equation}\label{eq:deepwalkreplacement_cdssm}
S \overset{EMB}{\rightarrow} \mathbb{R}^{n} \overset{MSE}{\rightarrow} \mathbb{R}^{m} \overset{CDSSM^{-1}}{\rightarrow} G \overset{\zeta^{-1}}{\rightarrow} \Omega 
\end{equation}

The CDSSM model consists of a three-layer convolutional neural network trained for creating embeddings of intent identifiers represented as sentences. In this work, we input to CDSSM the sequence of proto-taxonomies for each intent class. We refer to these algorithms as \emph{LSTM+C} and \emph{USE+C}, for EMB set with LSTM and USE, respectively.

\subsection{Out-of-Scope Sample Detection}
In this paper we are particularly interested to determine whether the proto-taxonomies improve the detection of out-of-scope (OOS) samples. A rejection mechanism based on a pre-defined threshold is used for OOS sample detection. This method can be easily applied to all of the methods described previously without the need neither for any specific training procedure nor OOS training data.

In greater detail, suppose that for each class $\omega_i \in \Omega$ there is a score denoted $\phi_i \in Z$, where $|Z| = |\Omega|$. Given that $\max(Z)$ represents the highest score associated to a class and that a rejection threshold $\theta$ has been defined on a validation set, samples can be classified as OOS whenever $\max(Z) < \theta$. If so, they are simply rejected, i.e., no classification output is produced for them. Otherwise, the sample is considered as an \emph{in-scope (IS)} sample and the classification is conducted normally.

The scores in $Z$ are represented either by the softmax probability for the traditional softmax-based methods or by the similarity of sentence and graph embeddings for the proposed approaches. For the latter, the similarity is computed by means of the dot product between those two embeddings.

\section{Metrics, Datasets, and Experiments}
In this section we present the experiments to evaluate the algorithms which use the proto-taxonomies with the neuro-symbolic algorithms described in the previous section. We explore their impact on the accuracy of intent recognition both in terms of classifying correctly utterances (in-scope accuracy) and determining which utterances are not covered by a set of intents (out-of-scope accuracy).

\subsection{Evaluation metrics} \label{sec:metrics}
We take into account a commonly-used metric for OOS dectection, i.e. \emph{equal error rate (EER)}~\cite{Tan2019} which corresponds to the classification error rate when the threshold $\theta$ is set to a value where \emph{false acceptance rate} (FAR) and \emph{false rejection rate} (FRR) are the closest. These two metrics are defined as:
\begin{equation} \small
    FAR = \frac{\mbox{Number of accepted OOS samples}}{\mbox{Total of OOS samples}}
\end{equation}
\begin{equation} \small
    FRR = \frac{\mbox{Number of rejected IS samples}}{\mbox{Total of IS samples}}
\end{equation}

In addition, \emph{in-scope error rate (ISER)} is considered to report IS performance, i.e. the accuracy considering only IS samples with $\theta$ set to zero, similar to the class error rate in \cite{Tan2019}. This metric is important to evaluate whether the alternative classification methods are able to keep up with the performance of their counterparts in the main classification task.

\subsection{The Larson and Telco Datasets}

 During the development and initial testing of the algorithms, we used two English datasets for in-depth experimentation. The first is the publicly-available \textit{Larson} dataset~\cite{larson-etal-2019-evaluation}; the second is a private real-world chatbot dataset used by telecommunications provider for customer care, called here the \textit{Telco} dataset. In the former, we added a proto-taxonomy by hand based on the identifiers of intents; in the latter case, we structured by hand original proto-taxonomy. The goal of the adjustments was to avoid spurious interference from taxonomy errors in the initial results.

In Larson there is a total of 22,500 in-scope samples, evenly distributed across 150 classes, where 18,000 examples are used for training and 4,500 for test. We conducted a simulation of OOS detection with the in-scope examples by doing five random samplings where we took out 30 intents and 3,600 training examples. We trained only with the remaining 120 intents and 14,400 examples. The test was then conducted on the 4,500 samples where 3,600 remained in-scope and 900 became OOS examples.

The Telco dataset contains 4,093 examples and 87 intents. From those, 3,069 examples were used for training and 1,024 for test. The OOS scenario was simulated by extracting different random samplings where 5 intents were removed. Given the smaller size of this dataset compared to Larson, we conducted 20 samplings instead of 5.

\begin{figure}[t!]
    \centering
    \includegraphics[trim=3cm 2cm 1cm 1cm,width=9cm]{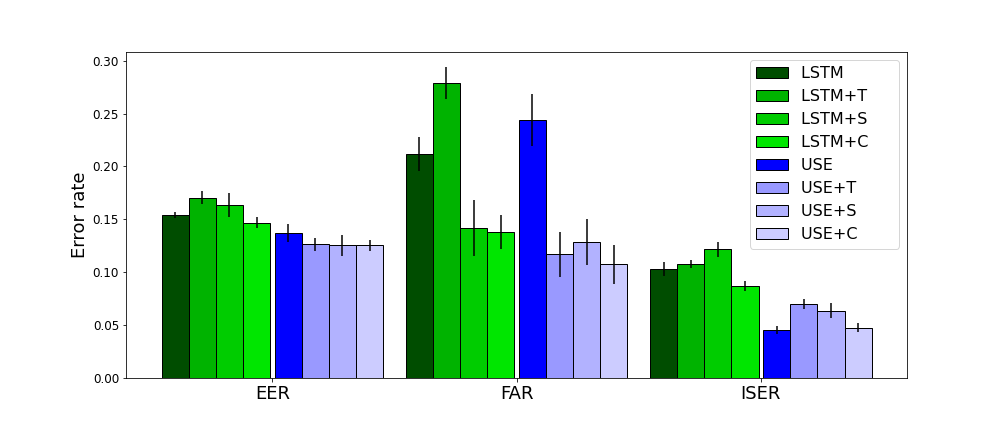}
    \caption{Different methods to incorporate the proto-taxonomy in Larson dataset, compared to the LSTM and USE baselines.}
    \label{fig:results_larson}
\end{figure}

\begin{figure}[t!]
    \centering
    \includegraphics[trim=3cm 2cm 1cm 1cm,width=9cm]{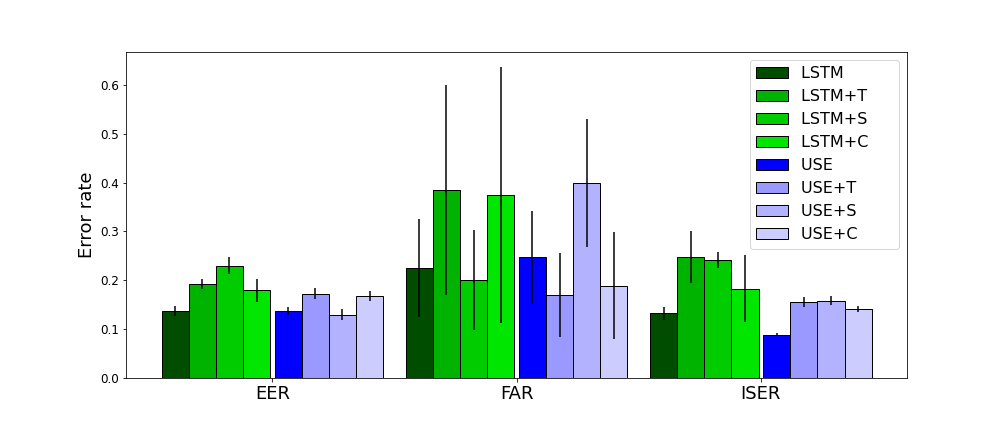}
    \caption{Different methods to incorporate the proto-taxonomy in Telco dataset, compared to the LSTM and USE baselines.}
    \label{fig:results_telco}
\end{figure}

For both sets we considered the following setup defined after preliminary evaluations. For the LSTM-based methods, the input sentence embedding size was set to 150 and output embeddings to 200. DeepWalk walk sizes were set to 20 for LSTM+T and USE+T. For both USE-based methods and the softmax-ones we trained a two-layer neural network with 800 hidden neurons. They were trained for 50 epochs.

\subsection{Results in the Larson and Telco Datasets}
The results on the Larson dataset are presented in fig.~\ref{fig:results_larson}. We observe that there can be a slight improvement in EER, in special with the USE-based and the LSTM+C methods. Nevertheless, there is a significant improvement in terms of FAR for all USE-based methods and LSTM+S and LSTM+C. Notice that even though the proposed approaches generally do not outperform LSTM and USE in ISER (except LSTM+C) the methods that approximate closer the ISER to the softmax counterparts tend to result in better EER and FAR rates.

In fig.~\ref{fig:results_telco}, the results on the Telco dataset show a different scenario. The proposed methods generally perform worse than or, at best, comparable to LSTM and USE in EER. In terms of FAR, some methods such as USE+T and USE+C seem to outperform but, considering the high standard deviation, the results are not significant. On the other hand, we also observe that the methods failed to get close in ISER compared with the softmax-based methods. That seems to indicate that for the cases where making use of meta-knowledge harms too much ISER, the symbolic knowledge creates noise and does not help improving either EER or FAR. 

There were two key findings from our experiments with the Larson and the Telco. First, the improvements using LSTM or USE as a base seem to be similar, possibly slightly better for the USE algorithm. Second, and most importantly, we saw much more improvements in the use of the proto-taxonomy in the Larson than in the Telco dataset, in spite of the similar nature of the datasets and the proto-taxonomies. This motivated us to try out the ideas in a larger and more diverse number of workspaces and solely focusing on USE to simplify the experiments.

\subsection{The ChatWorks Dataset}

We used the large set of real, professional workspaces from ChatWorks to create a dataset where our neuro-symbolic algorithms could be tested in a context of high diversity and realism. 
We started with the 3,840 workspaces available in English. 
To eliminate possible problems due to workspaces with poor quality, only workspaces with taxonomy rates over 30\% were considered. 
Next, workspaces with outliers in the number of intents and examples were removed following the $3\sigma$-rule, where values which extrapolate 3 standard deviations from the mean are not considered.
Finally, the ratio between the number of examples and intents must be greater than 10. From the filtered set we randomly selected 200 workspaces. 

The testing procedure involves the execution of 20 iterations for each workspace.
The tests are performed for all USE-based methods (USE, USE+T, USE+S, and USE+C).
Initially, the workspaces are split into training and test datasets (75\% and 25\%, respectively).
Next, the four methods are trained and tested on such datasets.
The evaluation metrics (EER, FAR and ISER) are measured on the results for the test datasets, being analyzed in terms of improvement of the proto-taxonomy models (USE+T, USE+S, and USE+C) in comparison to the base model (USE).

\subsection{Results in the ChatWorks Dataset}

Table~\ref{tab:chatworks_results} summarizes the results of the experiments with the ChatWorks dataset showing the distribution of the workspaces according to ranges of the percentage of improvement of each neuro-symbolic method compared to the baseline of USE (negative values signal worse than baseline). We highlight in boldface the best results for each range. Notice that when the neuro-symbolic method is worse than the baseline ($imp < -5\%$), smaller is better, and conversely for when it is better than the baseline ($5\% \leq imp$). The last column, \textit{best}, corresponds to the results considering the use of the best algorithm of the three algorithms.

\begin{table}[t!]
\centering
\footnotesize
\begin{tabular}{|c|r|r|r|r|}
\hline
\textbf{EER} & \multicolumn{4}{c|}{\% of workspaces} \\
\hline
improvement & USE+T & USE+S & \textbf{USE+C} & best\\ 
\hline
$imp < -5\%$ &        68\% & 60\% & \textbf{55\%} & 39\% \\
\hline
$-5\% \leq imp < 5\%$ &   16\% & 23\% & 18\% & 22\% \\
\hline
$5\% \leq imp < 10\%$ & 8\% & 8\% & \textbf{7\%}& 11\% \\
\hline
$10\% \leq imp < 20\%$ &   7\% & 7\% & \textbf{12\%} & \textbf{17\%} \\
\hline
$20\% \leq imp $ &        2\% & 4\% & \textbf{9\%} & \textbf{12\%} \\
\hline
\multicolumn{5}{c}{} \\
\hline
\textbf{FAR} & \multicolumn{4}{c|}{\% of workspaces} \\
\hline
improvement & USE+T & USE+S & \textbf{USE+C} & best\\ 
\hline
$imp < -5\%$ &        37\% & 47\% & \textbf{28\%} & 8\% \\
\hline
$-5\% \leq imp < 5\%$ &   17\% & 17\% & 17\% & 15\% \\
\hline
$5\% \leq imp < 10\%$ & 5\% & 7\% & \textbf{8\%} & 8\% \\
\hline
$10\% \leq imp < 20\%$ &   13\% & 11\% & \textbf{16\%} & \textbf{19\%} \\
\hline
$20\% \leq imp $ &        29\% & 19\% & \textbf{32\%} & \textbf{52\%} \\
\hline
\multicolumn{5}{c}{} \\
\hline
\textbf{ISER} & \multicolumn{4}{c|}{\% of workspaces} \\
\hline
improvement & USE+T & USE+S & \textbf{USE+C} & best\\ 
\hline
$imp < -5\%$ &        96\% & 95\% & \textbf{74\%} & 71\% \\
\hline
$-5\% \leq imp < 5\%$ &   4\% & 4\% & 20\% & 22\% \\
\hline
$5\% \leq imp < 10\%$ & 1\% & 1\% & \textbf{3\%} & 4\% \\
\hline
$10\% \leq imp < 20\%$ &   0\% & \textbf{1\%} & \textbf{1\%} & 1\% \\
\hline
$20\% \leq imp $ &        0\% & 1\% & \textbf{2\%} & 3\% \\
\hline
\end{tabular}
\caption{Percentage of workspaces on the ChatWorks dataset which saw different levels of improvement over the USE baseline, in terms of equal error rate (EER), false acceptance rate (FAR), and in-scope error rate (ISER).}
\label{tab:chatworks_results}
\end{table}

The results clearly indicate that the USE+C algorithm achieves the best results in all three metrics, although there is a significant portion of workspaces where the other methods are also competitive.
%, as we can conclude by some important differences from the USE+C percentages to the best algorithm shares. 
This is particularly true for the out-of-scope detection (FAR).

But, more important, the results support our claim that the meta-knowledge embedded by the developers can be used as input to neuro-symbolic algorithms to increase intent recognition performance. Notably in OOS detection, 71\% of the workspaces experienced an increase of 10\% or more in accuracy and more than 20\% of increase in 52\% of them. Even using only the best algorithm, USE+C, 48\% of the workspaces saw at least a 10\% improvement.

The overall results for accuracy, considering the EER metric, are also impressive. The top of table~\ref{tab:chatworks_results} shows that 28\% of the workspaces had improvements of 5\% or more with the USE+C algorithm and 29\% of them had an increase in 10\% or more in accuracy if they applied the best algorithm. However, the results for the in-scope accuracy (ISER) were much smaller with only about 6\% of the workspaces having an improvement of 5\% or more. We discuss these results and their implications next.

\section{Discussion and Future Work}
%Embedded Meta-Knowledge and Inverse Explainability}

We started this paper by showing evidence that there is a systematic practice of embedding symbolic knowledge into the intent identifiers among developers of professional conversational systems. We explored in detail the case of the ChatWorks platform and showed that a significant number of the workspaces have some sort of intent proto-taxonomy. This result was further validated in the workshops we had with professional ChatWorks developers.

The results of the experiments indicate that the intent proto-taxonomies embedded by those developers can indeed be used in many workspaces to improve accuracy in intent recognition. More than half of the workspaces drawn from ChatWorks saw improvements of more than 20\% in out-of-scope detection and in a little less than a third of them the overall error rate improved by 10\% or more. But why were there so many workspaces where we did not see impact?

First of all, we must take in account that the ChatWorks repository where we draw our dataset from has workspaces in different stages of development and deployment. We can expect significant differences in overall quality both in terms of intent definition and the utterance examples. We explored briefly basic characterizations of the proto-taxonomy quality, such as taxonomy rate, depth of the taxonomy, number of concepts, etc. but we saw no clear correlation with improvements in accuracy rates. We believe more complex metrics of knowledge structure need to be employed to better characterize which proto-taxonomies are good candidates. We plan to do so in our future work.

It is important to notice that, in the workspaces where we did see impact, the symbolic knowledge we mined was in an absolutely ``raw'' format. In spite of that, using the basic graph mining method described in the paper it was possible to obtain a ``meaningful'' knowledge structure, similar to a knowledge graph which could be used by our neuro-symbolic algorithms. To improve the quality of the taxonomies, we are working on designing an interface which allows the developers to manipulate directly the intent proto-taxonomy and, possibly, make it more correct, complete, and able to impact even more the intent recognition rates.

Moreover, we explored in this paper one particular case of symbolic knowledge embedding by developers of machine learning systems. However, it is unlikely that we will find in all machine learning development platforms similar patterns of knowledge embedding. We know, as discussed in the related work section, that people use similar proto-taxonomies when they name file and e-mail folders, when giving names to functions and variables in programs and data, and when writing comments into Jupyter notebooks. Also, platforms can foster further the use of meta-data and comments by developers, aiming to organically elicit usable knowledge, even if by doing so that knowledge is found to be inconsistent or incomplete. 

This can be a realistic path to knowledge acquisition for neuro-symbolic systems since we demonstrated here that such casual, organic, unsolicited knowledge can be mined and used effectively. It is likely that the fusion of neural and symbolic processing was key to handle the many mistakes and problems we found in that buried knowledge and we plan to explore this further in our experiments. But we hoped we have made the case that there is a hidden treasure of symbolic knowledge in many real-world systems and that robust neuro-symbolic methods such as the ones we described in the paper may be able to extract value from them.

\bibliographystyle{aaai}
\bibliography{bibliography}

\end{document}